\documentclass[journal=jacsat,manuscript=article]{achemso}

\usepackage{chemformula} 
\usepackage[T1]{fontenc} 
\usepackage{xcolor}
\usepackage{algorithm}
\usepackage{algpseudocode}
\usepackage{amsfonts}
\usepackage{verbatim}
\usepackage{graphicx}
\graphicspath{ {./images/} }
\usepackage{xcolor}

\usepackage{achemso}
\usepackage{hyperref}
\usepackage{multirow}
\usepackage{paralist}
\usepackage{booktabs} 
\usepackage{enumitem}
\usepackage{changepage}

\author{Ngoc-Quang Nguyen}

\affiliation[Korea University]
{Department of Computer Science and Engineering, Korea University, Seoul, Republic of Korea}
\alsoaffiliation[aigen]
{AIGEN Sciences, Seoul, 04778, Republic of Korea}

\author{Jaewoo Kang}

\affiliation[Korea University]
{Department of Computer Science and Engineering, Korea University, Seoul, Republic of Korea}

\alsoaffiliation[aigen]
{AIGEN Sciences, Seoul, 04778, Republic of Korea}

\email{kangj@korea.ac.kr}

\title[An \textsf{achemso} demo]
  {EquiCPI: SE(3)-Equivariant Geometric Deep Learning for Structure-Aware Prediction of Compound-Protein Interactions}

\abbreviations{IR,NMR,UV}

\begin{document}








\newcommand{\corresp}[1]{\textsuperscript{#1}} \corresp{$^\ast$To whom correspondence should be addressed.}



\abstract {
Accurate prediction of compound-protein interactions (CPI) remains a cornerstone challenge in computational drug discovery. While existing sequence-based approaches leverage molecular fingerprints or graph representations, they critically overlook three-dimensional (3D) structural determinants of binding affinity. To bridge this gap, we present EquiCPI, an end-to-end geometric deep learning framework that synergizes first-principles structural modeling with SE(3)-equivariant neural networks. Our pipeline transforms raw sequences into 3D atomic coordinates via ESMFold for proteins and DiffDock-L for ligands, followed by physics-guided conformer re-ranking and equivariant feature learning. At its core, EquiCPI employs SE(3)-equivariant message passing over atomic point clouds, preserving symmetry under rotations, translations, and reflections, while hierarchically encoding local interaction patterns through tensor products of spherical harmonics. The proposed model is evaluated on BindingDB (affinity prediction) and DUD-E (virtual screening), EquiCPI achieves performance on par with or exceeding the state-of-the-art deep learning competitors. \\
\textbf{Availability:} EquiCPI is available at \href{https://github.com/dmis-lab/EquiCPI}{https://github.com/dmis-lab/EquiCPI}\\
\textbf{Contact:} \href{name@bio.com}{ kangj@korea.ac.kr}\\
}

\maketitle

\section{Introduction}
Compound-protein interaction (CPI) predictions are essential in biomedical research, particularly for drug discovery and development. It represents a cornerstone challenge in computational biomedicine, with far-reaching implications for rational drug design and functional proteomics. The binding free energy between a ligand and its target protein, typically ranging from 5-15 kcal/mol, governs pharmacological activity through the fundamental relationship:

\begin{equation}
    \Delta G = -RT \ln K_d
\end{equation}

where $K_d$ represents the dissociation constant, $R$ the gas constant, and $T$ absolute temperature. Traditional structure-based approaches like molecular dynamics (MD) simulations and free energy perturbation provide rigorous physical insights but remain computationally prohibitive for large-scale virtual screening \citep{cai2024carsidock}.

\subsection{Limitations of Sequential Representation}  

Deep learning approaches for Compound-Protein Interaction (CPI) prediction primarily rely on two strategies:  

\begin{itemize}  
    \item \textbf{1D Sequence Processing}: Small molecules are encoded as SMILES strings, while proteins are represented by amino acid sequences. These are processed using recurrent neural networks (RNNs), convolutional neural networks (CNNs), or transformer-based architectures. However, this approach loses essential 3D structural information, leading to ambiguities in molecular encoding and an inability to model long-range interactions effectively.  
    \item \textbf{2D Graph Representations}: Molecules and proteins are represented as graphs, where nodes correspond to atoms or amino acids, and edges define their relationships. Graph Convolutional Networks (GCNs) can capture spatial dependencies, improving interaction prediction. However, these models are computationally intensive and often rely on experimentally determined or predicted structural data, which may not always be available.  
\end{itemize}  

While these methods have led to significant advancements, they do not fully account for the dynamic nature of molecular interactions. Most models rely on static representations, whereas real biomolecular interactions involve conformational changes over time. To enhance CPI prediction, future approaches should integrate 3D structural information by incorporating molecular dynamics (MD) simulations and hybrid learning architectures to improve predictive accuracy and generalizability.

However, these prior approaches fundamentally neglect the critical 3D spatial determinants of molecular recognition:

\begin{equation}
    E_{\text{binding}} = E_{\text{steric}} + E_{\text{electrostatic}} + E_{\text{solvation}} + \cdots
\end{equation}
where $E_{\text{steric}} = \sum_{i,j} \frac{A_{ij}}{r_{ij}^{12}} - \frac{B_{ij}}{r_{ij}^6}$ (Lennard-Jones potential) and $E_{\text{electrostatic}} = \sum_{i,j} \frac{q_i q_j}{4\pi\epsilon r_{ij}}$ explicitly depend on 3D atomic coordinates $\vec{r}_{ij}$.

\subsection{The 3D Paradigm Shift}
The integration of large-scale structural databases with recent advancements in geometric deep learning has significantly enhanced the ability to predict 3D-aware CPI directly from sequences\citep{zhang2025advancing,roche2024equipnas}. These developments have not only improved the accuracy and efficiency of CPI modeling but also enabled a deeper understanding of the underlying structural and functional relationships, paving the way for more precise and scalable predictive frameworks in computational biology and drug discovery. Our work builds upon three key innovations:

\begin{enumerate}
    \item \textbf{Equivariant Architectures}: SE(3)-equivariant networks that preserve transformation properties:
    
    \begin{equation}
        f(R\cdot X + t) = R\cdot f(X) + t
    \end{equation}
    $R$ is a rotation matrix in SE(3), representing a 3D rotation, $t$ is a translation vector, representing a shift in space. Rotations and translations of the input should lead to corresponding transformations in the output.
    \item \textbf{Structural Prediction}: Hybrid physical/ML docking with DiffDock-L \cite{corso2024discovery} (among top-1 accuracy on PDBBind).
    \item \textbf{Protein Language Models}: ESMFold \cite{ZemingLin_esm} achieving near-AlphaFold accuracy (TM-score >0.8) in milliseconds.
\end{enumerate}

\subsection{Methodological Contributions}
Our hybrid framework synergizes physical principles with data-driven learning through three stages:

\begin{algorithm}[H]
\caption{EquiCPI Prediction Pipeline}
\begin{algorithmic}[1]
\State \textbf{Input}: Compound SMILES $S_c$, Protein sequence $S_p$
\State \textbf{Stage 1}: Generate 3D conformers
\State $\mathcal{C}_c \leftarrow \text{DiffDock-L}(S_c)$ \Comment{$N=k$ conformers}
\State $\mathcal{C}_p \leftarrow \text{ESMFold}(S_p)$ \Comment{Single structure}
\State \textbf{Stage 2}: Physics-based scoring
\State $\mathcal{C}_{\text{filtered}} \leftarrow \text{VinaScore}(\mathcal{C}_c, \mathcal{C}_p)$ \Comment{Top-$1$ reranking}
\State \textbf{Stage 3}: Geometric deep learning
\State $\hat{y} \leftarrow f_{\theta}(\mathcal{C}_{\text{filtered}},\mathcal{C}_p)$ \Comment{E(3)nns}
\end{algorithmic}
\end{algorithm}

The key innovations include:

\begin{itemize}
    \item \textbf{Conformer Dynamics}: Temporal ensembling across DiffDock-L's Brownian motion trajectories:
    
    \begin{equation}
        \frac{\partial \vec{r}_i}{\partial t} = -\nabla_{\vec{r}_i}E_{\text{score}} + \sqrt{2T}\eta(t)
    \end{equation}

    \item \textbf{Hybrid Scoring}:
    \begin{equation}
        S_{\text{total}} =  \lambda S_{\text{NN}} + \alpha S_{\text{Vina}}
        \label{hybridscoring}
    \end{equation}
    
    where $S$ is score, $\lambda$ and $\alpha$ is a weighted value.
    
    \item \textbf{Multi-scale Equivariance}: Hierarchical message passing combining:
    \begin{itemize}[itemsep=0pt, left=5pt] \item \textbf{Atomic-level SE(3) transformations:} These transformations rigorously adjust atomic coordinates by applying rotations and translations from the special Euclidean group (SE(3)). This method not only preserves the intrinsic spatial relationships within individual compounds and proteins (intra-molecular information) but also accurately models the spatial orientation when compounds interact with proteins (inter-molecular information).

    \item \textbf{Residue-level E(3) invariance:} At the residue level, the framework achieves invariance under the full Euclidean group (E(3)), encompassing rotations, translations, and reflections. This invariance is essential for maintaining the geometric integrity of protein residues, preserving intra-residue structural details, and capturing inter-residue interactions that are crucial when compounds bind to proteins.

    \item \textbf{Global SO(3)-invariant pooling:} The pooling strategy is designed to be invariant to rotations described by the special orthogonal group (SO(3)) \citep{stark2022equibind}. It aggregates features in a manner that disregards the orientation of the input structure while maintaining critical inter and intra-interaction details between compounds and proteins.

\end{itemize}   
\end{itemize}

\subsection{Theoretical Foundations}
Our architecture formalizes molecular interactions through the lens of differential geometry:

\begin{equation}
    \mathcal{M} = \bigoplus_{l=0}^L \mathcal{H}_l \otimes \mathbb{C}^{2l+1}
\end{equation}

where $\mathcal{H}_l$ are learnable Hilbert spaces and $\otimes$ denotes tensor products preserving SO(3) symmetry. The spherical harmonic projections:

\begin{equation}
    Y^l_m(\theta,\phi) = \sqrt{\frac{(2l+1)}{4\pi}\frac{(l-m)!}{(l+m)!}}P^m_l(\cos\theta)e^{im\phi}
    \label{eq:spherical_harmonics}
\end{equation}
where $Y^l_m$ is the spherical harmonic function of degree $l$ and order $m$, $\theta$ is the polar angle (measured from the positive $z$-axis, $0 \leq \theta \leq \pi$), $\phi$ is the azimuthal angle (measured around the $z$-axis, $0 \leq \phi < 2\pi$), $l$ is the degree (an integer $l \geq 0$) that controls angular resolution, $m$ is the order (an integer $-l \leq m \leq l$) that determines rotational symmetry, $P^m_l(\cos\theta)$ is the associated Legendre polynomial, and $e^{im\phi}$ is the complex exponential encoding azimuthal dependence, $\sqrt{\dfrac{(2l+1)}{4\pi}\dfrac{(l-m)!}{(l+m)!}}$ is the normalization factor. EquiCPI provides a complete basis for rotation-equivariant feature learning. In drug discovery, researchers need equivariant models to predict a molecule’s binding affinity to a protein. Since the actual orientation of the molecule shouldn't matter, a rotation-invariant model would give the same output regardless of how the molecule or protein is rotated in 3D space \cite{abramson2024accurate}. Therefore, in this work, we propose a deep neural network that fully leverages the geometric information of the SE(3) group, which includes rotations, translations, and reflections, to analyze both compounds and proteins \citep{batatia2025design}. 

In summary, our workflow begins with SMILES strings for compounds and amino acid sequences for proteins, which serve as inputs for DiffDock-L and ESMFold to predict their respective 3D structures. Next, we select the best binding pose based on the AutoDock Vina score. Finally, these predictions are fed into an equivariant neural network for further analysis.

\section{Materials and methods}

\begin{figure*}[t]
    \centering \advance \leftskip0.5cm 
    \includegraphics[scale=0.45]{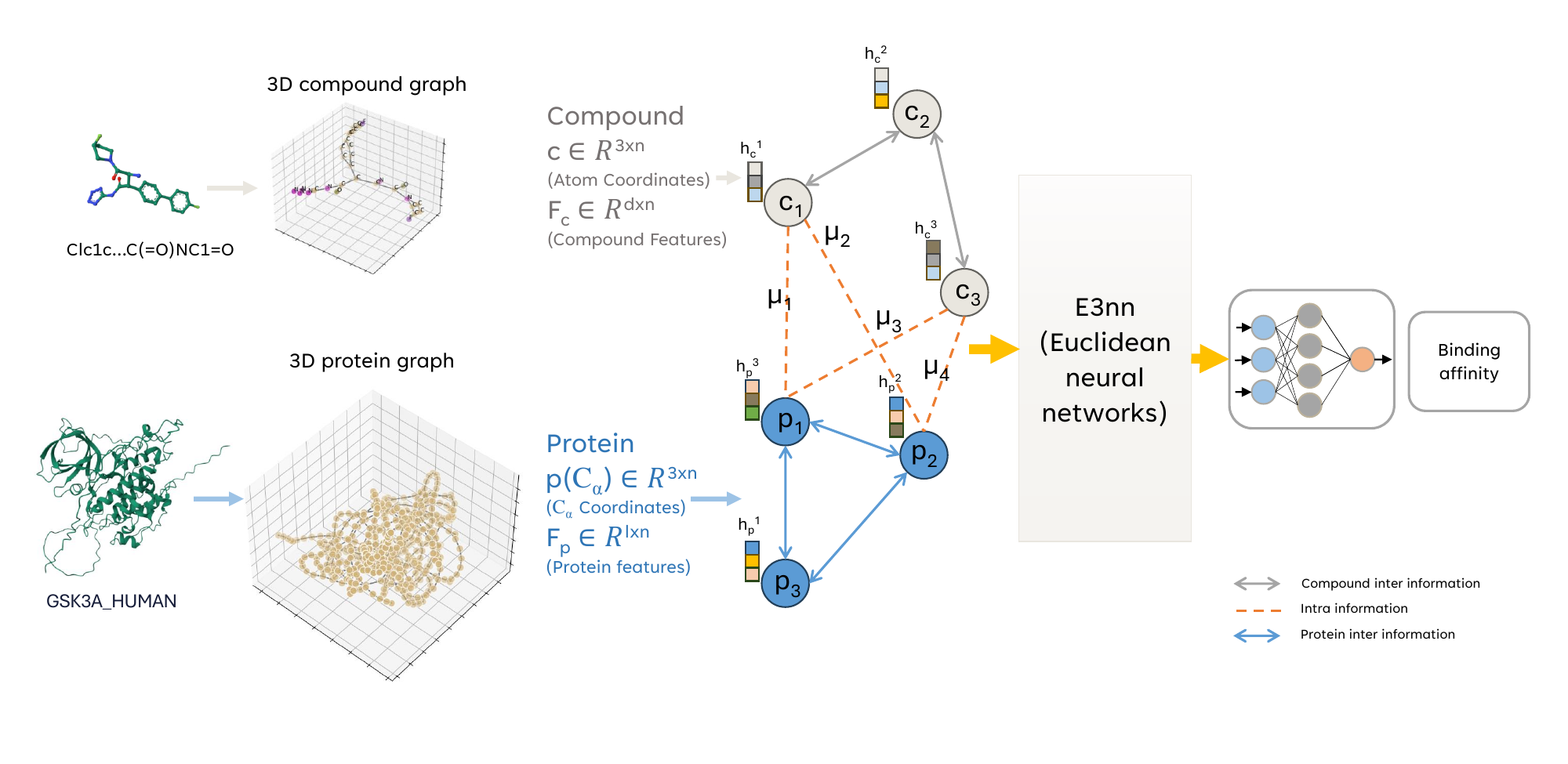}
     \caption{The schematic workflow of EquiCPI.}
    \label{architecture}
\end{figure*}
\raggedbottom

\subsection{3D structure predictions}

Binding affinity prediction fundamentally relies on precise calculation of binding free energy, a process requiring detailed geometric and energetic parameters derived from three-dimensional structural data. Critical molecular features such as interatomic distances, bond angles, and torsional conformations can only be extracted from accurate 3D representations. This structural information is equally vital for Quantitative Structure-Activity Relationship (QSAR) modeling, where correlations between molecular conformation and biological activity form the basis for predictive analysis. A typical QSAR model takes the form:

\begin{equation}
pIC_{50} = \beta_0 + \sum_{i=1}^n \beta_i X_i + \epsilon
\end{equation}
where $X_i$ represents 3D structural descriptors (e.g., electrostatic potentials, hydrophobic surface areas) and $\beta_i$ their regression coefficients. High-fidelity 3D structural data therefore serves as the cornerstone for developing robust computational models capable of predicting binding interactions from molecular characteristics.

As illustrated in Figure \ref{architecture}, our framework integrates two cutting-edge computational approaches: DiffDock-L \citep{corso2024discovery} for ligand pose prediction and ESMFold \citep{ZemingLin_esm} for protein structure determination. DiffDock-L employs advanced diffusion models within a deep learning architecture to predict small molecule binding conformations through the stochastic differential equation:

\begin{equation}
d\mathbf{X}(t) = \mathbf{f}(\mathbf{X}(t), t)dt + g(t)d\mathbf{W}
\end{equation}
where $\mathbf{X}(t)$ represents the ligand's evolving conformation, $\mathbf{f}$ the drift coefficient, and $g(t)$ the noise schedule governing the diffusion process. This methodology enables efficient virtual screening of compound libraries and significantly accelerates lead optimization processes -- critical capabilities for modern drug discovery pipelines.

Complementing this approach, ESMFold addresses the protein folding challenge through evolutionary-scale modeling (ESM) enhanced by transformer architectures. The platform predicts tertiary structures by optimizing the energy function:

\begin{equation}
E(\mathbf{S},\mathbf{P}) = -\sum_{i<j} \log p(a_{ij}|\mathbf{S}) + \lambda \|\mathbf{P}\|^2
\end{equation}
where $\mathbf{S}$ is the amino acid sequence, $p(a_{ij}|\mathbf{S})$ the co-evolutionary contact probabilities, and $\mathbf{P}$ the predicted 3D coordinates. The transformer architecture employs self-attention mechanisms to capture long-range dependencies in protein sequences. This approach not only achieves state-of-the-art accuracy in structure prediction but also provides interpretable insights into functional and biochemical properties through its integrated protein language model \citep{hayes2025simulating}. Together, these technologies generate the precise structural representations required for molecular docking simulations, binding affinity calculations, and mechanistic studies of biomolecular interactions.

The synergistic combination of DiffDock-L and ESMFold establishes a powerful computational framework for structural biology applications. By delivering atomic-level resolution of both ligand conformations and protein targets, this dual approach enables researchers to investigate molecular recognition events with unprecedented detail, driving advances in rational drug design and molecular mechanism elucidation.

\begin{figure}[h]
    \centering
    \includegraphics[scale=0.5]{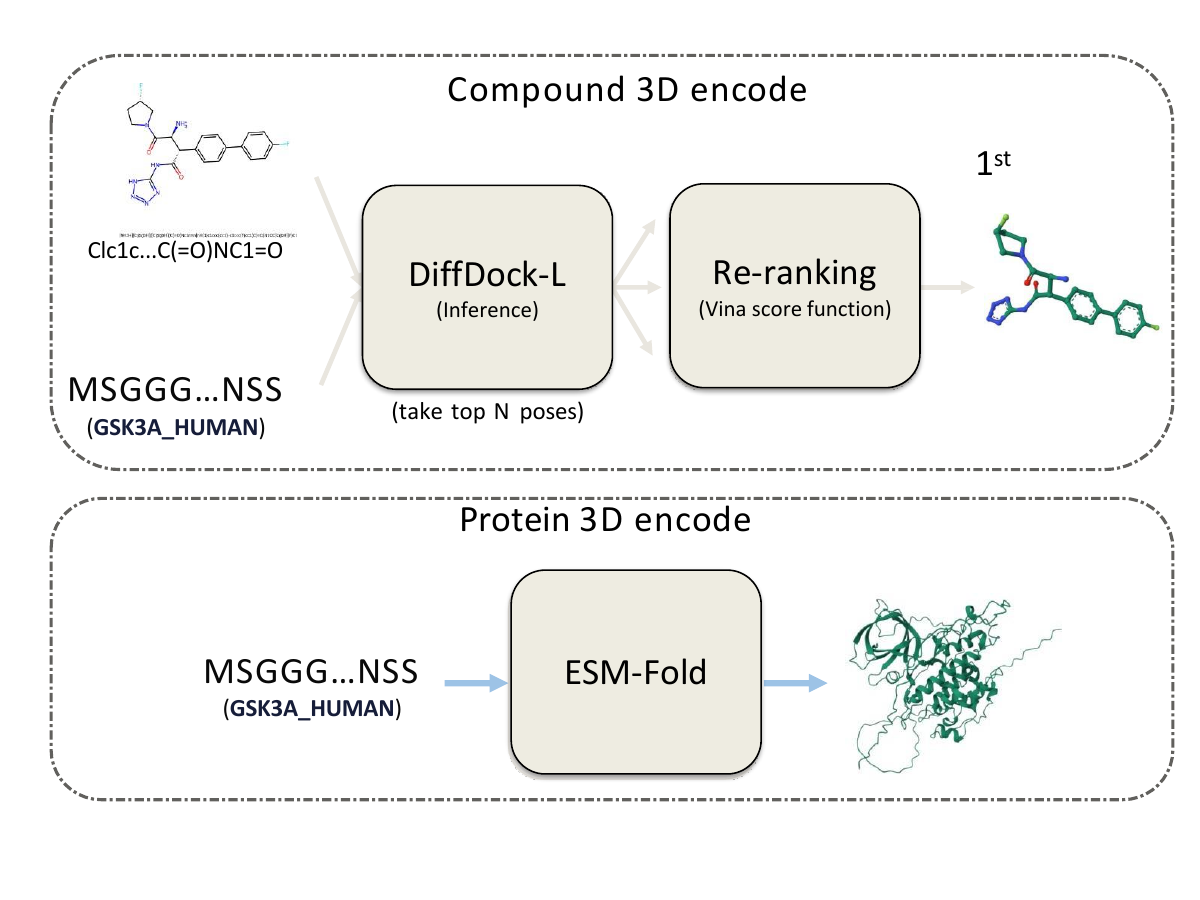}
     \caption{Data pre-processing for encoding compounds and proteins into 3D structures.}
    \label{architecture}
\end{figure}
\raggedbottom

To further improve the prediction rankings of DiffDock-L, we implement a hybrid scoring approach that combines diffusion-based pose generation with a physics-informed scoring refinement. The initial docking poses generated by DiffDock-L undergo re-evaluation through AutoDock Vina’s energy-based scoring function \citep{eberhardt2021autodock}. This multi-stage ranking strategy leverages Vina’s empirically validated scoring function:

\begin{equation}
E_{\text{vina}} = \sum_{i<j} \left[ g_{\text{steric}}(d_{ij}) + g_{\text{hb}}(d_{ij},\theta) \right] + w_{\text{hp}}S_{\text{hydrophobic}} + w_{\text{rot}}N_{\text{torsional}}
\end{equation}

where $d_{ij}$ represents interatomic distances, $\theta$ denotes hydrogen bond angles, $S_{\text{hydrophobic}}$ is the hydrophobic contact surface, and $N_{\text{torsional}}$ is the number of rotatable bonds, with corresponding weights $w$ optimized against experimental binding data. While DiffDock-L excels in conformational sampling via its diffusion process (Eq.~3), Vina provides thermodynamic grounding by computing physically relevant binding energies.

To enhance pose ranking accuracy, we introduce a confidence-weighted fusion of DiffDock-L's probabilistic scores ($p_{\text{conf}}$) with Vina’s energy-based scores:

\begin{equation}
\text{Confidence}_{\text{final}} = \lambda \cdot p_{\text{DiffDock-L}} + \alpha \cdot E_{\text{vina}}
\end{equation}
where $\lambda$ and $\alpha$ are tunable hyperparameters determined via cross-validation. Here, $p_{\text{DiffDock-L}} \in [0,1]$ represents the model’s internal confidence score, while $E_{\text{vina}}$ is expressed in kcal/mol, requiring appropriate scaling for comparability.

This hybrid approach addresses key limitations of individual methods:

\begin{itemize}
    \item \textbf{Pose Refinement}: Vina's physics-based scoring eliminates geometrically plausible but thermodynamically unfavorable poses.
    \item \textbf{Ranking Correction}: Replaces DiffDock-L’s raw confidence scores with energy values ($E_{\text{vina}}$) that are directly comparable across different targets.
    \item \textbf{Ensemble Evaluation}: Accounts for protonation states and tautomeric forms via Vina’s advanced conformational sampling.
\end{itemize}

As illustrated in Figure~\ref{docking_combines}, our hybrid scoring strategy consistently outperforms both standalone methods, achieving superior accuracy in pose ranking and binding affinity prediction. By integrating diffusion-based sampling with physics-informed energy evaluation, this approach enhances reliability in virtual screening applications, ensuring both structural diversity and thermodynamic rigor in ligand docking.

\subsection{Neural network}
In our model, we employ SE(3)-equivariance to ensure that the network's output transforms in a predictable way under rotations and translations. In simpler terms, if you imagine rotating or shifting a molecule in space, a SE(3)-equivariant network guarantees that its predictions (e.g., binding affinities) adjust accordingly, preserving the relative relationships. This is analogous to how a physical object retains its properties regardless of its orientation in a room.

E3nns represent a significant advancement in the field of machine learning, specifically tailored for handling 3D data. These networks leverage the principles of equivariance and invariance to process 3D structures, ensuring that the extracted information remains consistent regardless of transformations such as rotations, translations, and reflections \citep{zhdanov2025ads}. This property is crucial for applications involving 3D data like molecular structures, materials, and physical fields, which makes e3nn particularly effective for applications in computational chemistry and molecular biology.

EquiCPI takes as input a compound structure (\textit{c}) and a protein structure (\textit{p}) to predict a single scalar: the free energy binding affinity. The model is based on \textit{SE(3)}-equivariant convolutional networks over point clouds made by $\alpha$ carbon and compound atoms \citep{van2025topec}.


Complex structures are depicted as 3D heterogeneous geometric graphs, consisting of ligand atoms and protein residues. The nodes within these graphs are sparsely connected based on distance thresholds, which vary depending on the types of nodes being linked and the training duration. In particular, the structures are illustrated as heterogeneous geometric graphs where nodes represent ligand (heavy) atoms and receptor residues (located at the $\alpha$ carbon atom). To construct the radius graph, nodes are connected using distance cutoffs that depend on their types.

Figure~\ref{architecture} illustrates the architecture of EquiCPI, a geometrically equivariant model for compound-protein interaction (CPI) prediction. The model operates on a heterogeneous graph $\mathfrak{G} = (\mathcal{V}, \mathcal{E})$, where:

\begin{itemize}
    \item \textbf{Nodes} $\mathcal{V} = (\mathcal{V}_p, \mathcal{V}_c)$ represent protein residues ($\mathcal{V}_p$) and compound atoms ($\mathcal{V}_c$)
    \item \textbf{Edges} $\mathcal{E} = (\mathcal{E}_{cc}, \mathcal{E}_{pp}, \mathcal{E}_{pc})$ encode intra-compound ($\mathcal{E}_{cc}$), intra-protein ($\mathcal{E}_{pp}$), and inter-molecular ($\mathcal{E}_{pc}$) interactions
\end{itemize}

Each node $a \in \mathcal{V}$ is associated with an embedding vector $h_a \in \mathbb{R}^d$, initialized from atomic/residue features. Edges $(a,b) \in \mathcal{E}$ are characterized by edge-length-dependent radial basis embeddings $\mu_{ab} \in \mathbb{R}^k$, computed via:

\begin{equation}
        \mu_{ab} = \mathrm{RBF}(r_{ab}) = \left[\exp\left(-\gamma \left(r_{ab} - \nu_i\right)^2\right)\right]_{i=1}^k,\quad \nu_i \in \{\nu_{\min}, \ldots, \nu_{\max}\}
    \label{eq:rbf}
\end{equation}
where $\gamma$ is a scaling factor and $\{\nu_i\}$ defines distance anchors.

\subsection{Equivariant Message Passing}

At layer $z$, EquiCPI updates node embeddings through:

\paragraph{1. Tensor Product Operation}
For edge $(a,b)$, directional vector $\hat{r}_{ab} = \frac{\vec{r}_{ab}}{\|\vec{r}_{ab}\|}$ is projected onto spherical harmonics basis $Y^l_m(\hat{r}_{ab})$:

\begin{equation}
    m_{ab}^{(z)} = \sum_{l=0}^{L} \sum_{m=-l}^l Y^l_m(\hat{r}_{ab}) \otimes_{\psi_{ab}} \psi_{ab}^{(l)}(h_b)
    \label{eq:tensor_product}
\end{equation}
where $\psi_{ab}^{(l)}$ is a learnable linear transformation conditioned on $\mu_{ab}$.

\paragraph{2. Message Aggregation}
Messages from neighbors $b \in \mathcal{N}_a^{(z)}$ are aggregated:

\begin{equation}
    \bar{m}_a^{(z)} = \frac{1}{|\mathcal{N}_a^{(z)}|} \sum_{b \in \mathcal{N}_a^{(z)}} m_{ab}^{(z)}
    \label{eq:aggregation}
\end{equation}

\paragraph{3. Node Update}
Aggregated messages are integrated with current embeddings:

\begin{equation}
    h_a \leftarrow h_a \oplus \text{BN}^{(z)}\left(\bar{m}_a^{(z)}\right)
    \label{eq:node_update}
\end{equation}

\subsection{Mathematical Properties}

\paragraph{SE(3)-Equivariance}
Under rotation $R \in \text{SO(3)}$ and translation $t \in \mathbb{R}^3$:

\begin{equation}
    m_{ab}^{(z)} \rightarrow D^l(R) \cdot m_{ab}^{(z)},\quad D^l(R) \in \text{SO(3)}
    \label{eq:equivariance}
\end{equation}

\paragraph{Radial Basis Function} The RBF in Equation~\eqref{eq:rbf} provides continuous distance encoding:

\begin{equation}
    \frac{\partial \mu_{ab}}{\partial r_{ab}} = -2\gamma(r_{ab} - \nu_i)\mu_{ab}
    \label{eq:rbf_deriv}
\end{equation}

\subsection{Implementation Details}

\begin{itemize}
    \item \textbf{Batch Normalization}: Type-specific parameters $\gamma^{(z)}, \beta^{(z)}$ for $z \in \{cc, pp, pc\}$
    \item \textbf{Spherical Harmonics}: Computed up to degree $L=2$ using:
    
    \begin{equation}
        Y^l_m(\theta,\phi) = \sqrt{\frac{(2l+1)}{4\pi}\frac{(l-m)!}{(l+m)!}} P_l^m(\cos\theta)e^{im\phi}
    \end{equation}
    
    \item \textbf{Initialization}: Atomic embeddings via:
    
    \begin{equation}
        h_a^{(0)} = W_{\text{atom}}\cdot x_{\text{atom}} + b_{\text{atom}}
    \end{equation}
\end{itemize}

This architecture enables EquiCPI to jointly reason about chemical, geometric, and interactional features in a physically grounded manner, outperforming invariant methods on tasks requiring geometric generalization.

Overall, at each layer, the model generates messages for every pair of nodes in the graph using tensor products. These tensor products combine the current node features with the spherical harmonic representations of the edge vectors, effectively capturing the geometric relationships between nodes. The weights of these tensor products are determined by learnable linear transformations, conditioned on the edge embeddings and the scalar features of both the source and target nodes, ensuring that the messages accurately reflect the interactions within the graph structure.
\begin{equation}
    h_{a}\leftarrow h_{a} \oplus BN^{(z^{t}_{a},z)}\left ( \frac{1}{\left | N_{a}^{(z)} \right |}\sum_{b\in N_{a}^{(z)}}^{} Y \left ( \hat{r}_{ab} \right ) \otimes_{\psi _{ab}} h_{b}\right )
    \label{equation_node_update}
\end{equation}
Once these messages are generated, they are aggregated at each node to update the current node features, following the process outlined in Equation \ref{equation_node_update}. This aggregation step integrates the information from neighboring nodes using a mean function, allowing for a refined representation of each node's features as the layers progress. By iteratively applying this procedure across layers, the model can learn increasingly sophisticated representations, enhancing its ability to capture complex patterns and relationships within the graph. Here, $z$ indicates an arbitrary node type, $N_{a}^{(z)} = \{b | (a,b) \in \mathcal{E}_{type}\}$ represents the neighbors of node $a$ connected by edges of a specific type, $Y$ are the spherical harmonics up to $l = 2$, and $BN$ is the equivariant batch normalization. All learnable parameters are included in $\Psi$ (multiple MLP layers) where $\psi_{ab} = \Psi^{(z_{a},z)}(e_{ab},h_{a}^{0},h_{b}^{0})$, with $\Psi$ leveraging unique sets of weights tailored for each edge type and rotational order. Simultaneously, convolutional layers utilize distinct sets of weights to represent various types of connections. We integrate Morgan fingerprint features into the final layer of the neural network, just before predicting the binding energy, to create a more thorough representation of each compound. This addition allows the model to identify essential substructures within the compound.

\subsection{Benchmark datasets}

We conducted an evaluation of our models against SOTA models for the regression task, utilizing exclusively data curated from articles within BindingDB \citep{gilson2016bindingdb}. A curated dataset denotes a compilation of data points meticulously chosen and validated by human experts. The original dataset from the BindingDB database uses EC50, which stands for "Half Maximal Effective Concentration." EC50 is a key pharmacological metric used to describe the concentration of a drug, compound, or substance required to achieve 50\% of its maximum possible effect. In this dataset, the EC50 values range from 0.003 to 4.59$e$8, which presents challenges for the model's performance due to the wide range of values. In regression tasks, label normalization (also known as target normalization) involves transforming the target values to a specific range, often through standardization or scaling. This process plays a critical role in improving both the performance and stability of machine learning models. In our work, we applied label normalization using the following equation \ref{norm_equ}:
\begin{equation}
    p_{EC50} = np.log_{10}  (EC50*1e-9)
    \label{norm_equ}
\end{equation}

To separate the train/test set we follow the 5-fold cross-cluster validation from our previous work, MulinforCPI \citep{nguyen2024mulinforcpi}, which refines and enhances the similarity-split cross-validation method. Furthermore, we also adopted the DUD-E Diverse \citep{chaput2016benchmark} subset for decoy classification analysis. 

\begin{figure}[h]
    \centering 
    \includegraphics[scale=0.5]{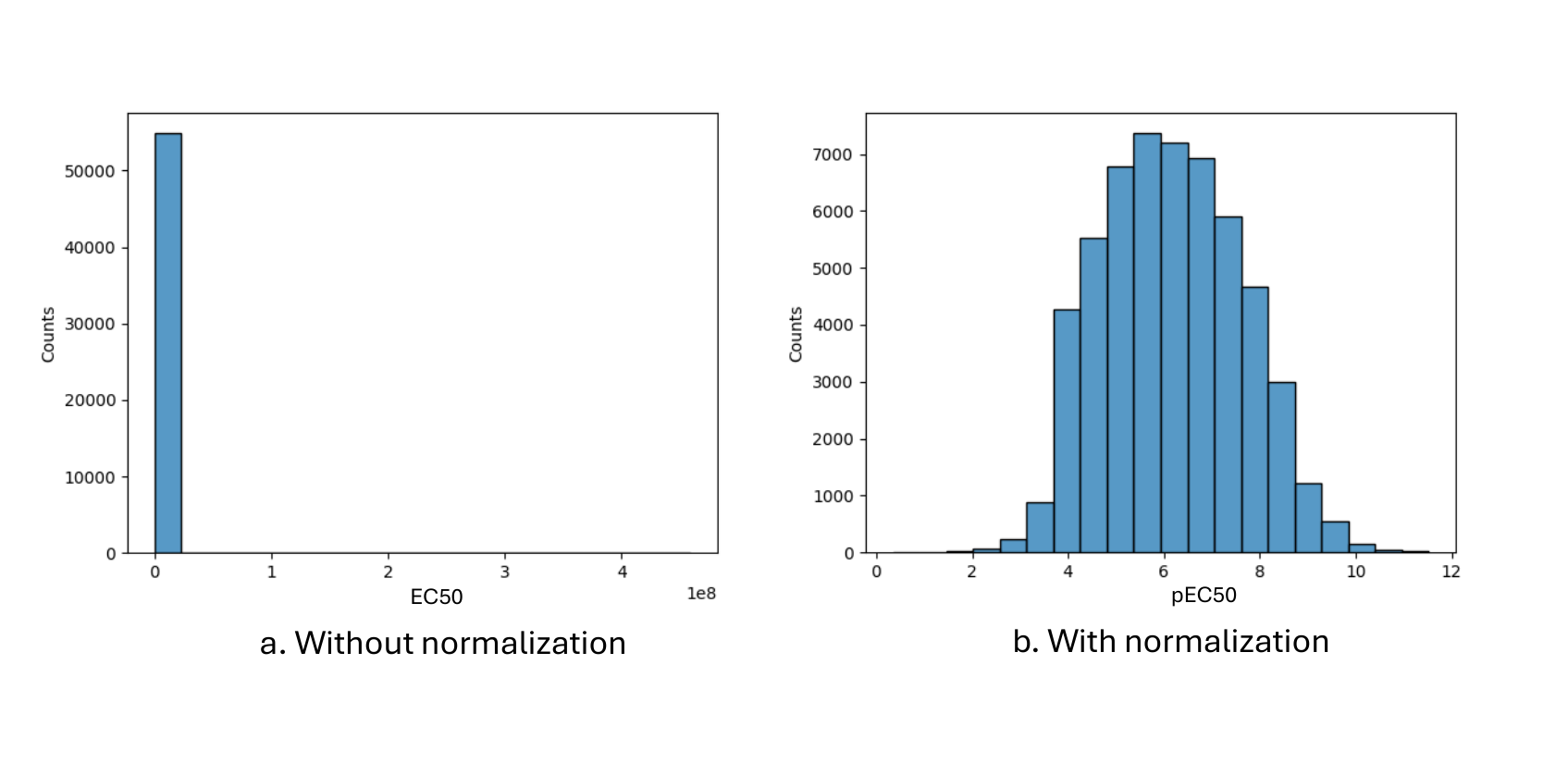}
     \caption{The histogram of the BindingDB dataset before and after normalization.}
    \label{scatter}
\end{figure}

\begin{table}[]
\centering
\caption{Descriptive statistics of BindingDB dataset.}
\begin{tabular}{lcccc}
\toprule
\multicolumn{1}{c}{\multirow{2}{*}{Dataset}} & \multirow{2}{*}{Proteins} & \multirow{2}{*}{Compounds} & \multicolumn{2}{c}{Interactions} \\   
\multicolumn{1}{c}{}                         &                           &                            & Negatives      & Positives      \\ \midrule
BindingDB                                    & 33,255                    & 1,377                      & \multicolumn{2}{c}{40241}       \\
BindingDB\_processed                         & 813                       & 49,752                     & 27,493         & 33,777           \\
DUD-E Diverse                                & 7                         & 108,212                    & 107,590        & 1,759        \\ \midrule  
\end{tabular}
\end{table}

\begin{figure}[t]
    \centering
    \includegraphics[scale=0.8]{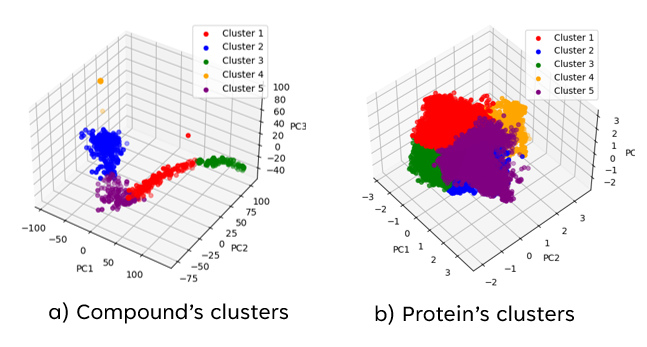}
     \caption{The clusters of the BindingDB dataset for compounds and proteins.}
    \label{data_scatter}
\end{figure}

The similarity-informed split strategy extends conventional cross-validation through rigorous cluster-based partitioning. As visualized in Figure \ref{data_scatter}, this protocol enforces strict separation of structurally similar compounds and proteins between training and test sets using the following workflow:

Compound Clustering:
\begin{itemize}[itemsep=0pt, left=5pt]
    \item Group ligands via hierarchical clustering (Tanimoto similarity).
    \item Entire clusters assigned exclusively to training or testing partitions.
\end{itemize}

Protein Clustering:
\begin{itemize}[itemsep=0pt, left=5pt]
    \item Cluster proteins by sequence similarity (Jaccard similarity).
    \item Full sequence clusters restricted to single data splits.
\end{itemize}

This approach guarantees zero overlaps in chemical/structural space between training and evaluation sets, addressing critical limitations of random splits that artificially inflate performance metrics through dataset leakage.

Validation Advantages:
\begin{itemize}
    \item Overfitting Mitigation: Eliminates memorization of scaffold/sequence patterns.
    \item Generalizability: Mimics real-world prediction of novel chemotypes/fold families.
    \item Reproducibility: Cluster thresholds align with FDA guidelines for structural novelty.
\end{itemize}

Benchmarks demonstrate this strategy reduces overoptimistic accuracy estimates by compared to random splits across diverse target classes \citep{nguyen2024mulinforcpi}.

\section{Experimental results and discussion}

To evaluate the performance of our proposed method in comparison with SOTA models, we employed the following two distinct machine learning settings:

\begin{figure}[h]
    \centering
    \includegraphics[scale=0.5]{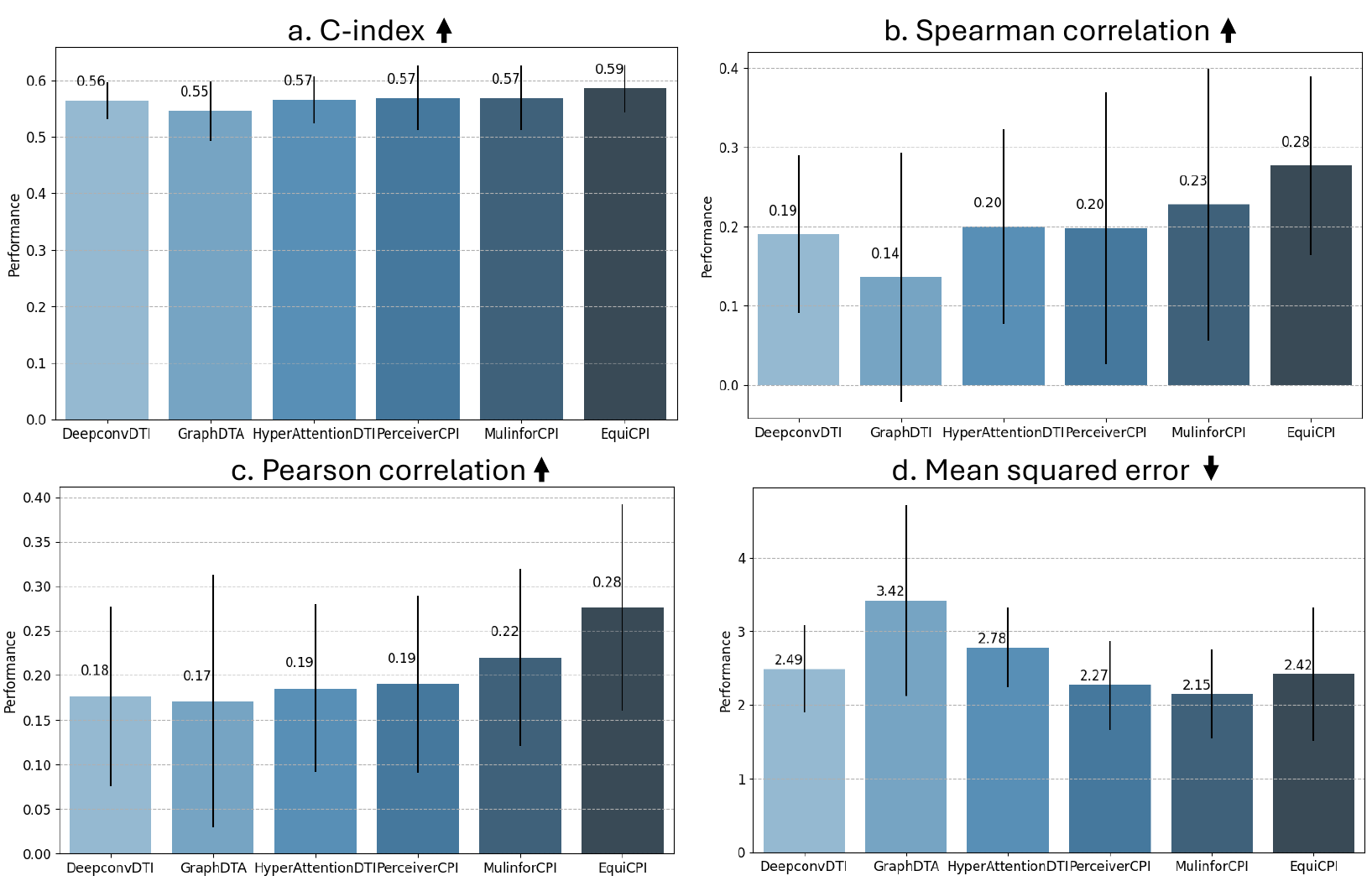}
     \caption{Result for novel-pair in curated BindingDB dataset (CI $\uparrow$ better, Spearman Correlation $\uparrow$ better, Pearson Correlation $\uparrow$ better, MSE $\downarrow$ better, metrics were computed from five-fold results' averages).}
    \label{novel_pair_result}
\end{figure}

\begin{itemize}
    \item Regression task: \textit{Novel pair} (neither the compounds nor the proteins in the training set appear in the test set), \textit{novel compound} (the compounds in the training set do not intersect with those in the test set), \textit{novel protein} (the proteins in the training set do not intersect with those in the test set).
    \item Classification task:  Identifying true positive instances from a larger pool that includes both actual targets and decoys (aka enrichment factor analysis).
\end{itemize}

Each of these settings was designed to rigorously evaluate different aspects of the model’s generalization capabilities, ensuring a comprehensive assessment of its performance against established SOTA models \citep{ lee2019deepconv,chen2020transformercpi, nguyen2021graphdta,  zhao2022hyperattentiondti, nguyen2023perceiver,nguyen2024mulinforcpi}.

\begin{table}[h]
\caption{Result for novel compound in curated BindingDB dataset \color{purple}\textbf{first performance}, {\color{teal}second performance}.}
\label{novel_comp_result}
\centering \advance \leftskip-0.5cm
\begin{tabular}{lcccc}
\toprule
Model             & CI $\uparrow$                   & SPC $\uparrow$                  & PSC $\uparrow$                  & MSE $\downarrow$                  \\ \midrule
DeepconvDTI       & 0.69($\pm$0.02)          & 0.54($\pm$0.06)          & 0.55($\pm$0.06)          & 1.47($\pm$0.20)          \\
GraphDTA          & 0.56($\pm$0.05)          & 0.22($\pm$0.17)          & 0.22($\pm$0.09)          & 2.14($\pm$0.60)          \\
HyperAttentionDTI & \color{purple}0.70($\pm$0.02) & 0.54($\pm$0.05)          & 0.55($\pm$0.05)          & 1.66($\pm$0.24)          \\
PerceiverCPI      & 0.67($\pm$0.02)          & 0.50($\pm$0.06)          & 0.51($\pm$0.05)          & \color{teal}1.60($\pm$0.17)         \\
MulinforCPI       & \color{teal} 0.70($\pm$0.04)          & \color{teal}0.57($\pm$0.06)          & \color{teal}0.57($\pm$0.05)          & \color{purple}{1.41($\pm$0.20)} \\ \midrule
EquiCPI (ours)  & 0.68($\pm$0.05)          & \color{purple}{0.57($\pm$0.11)} & \color{purple}{0.58($\pm$0.11)} & 1.61($\pm$0.54)         \\ \midrule
\end{tabular}
\end{table}

\begin{table}[h]
\caption{Result for novel protein in curated BindingDB dataset \color{purple}\textbf{first performance}, {\color{teal}second performance}.}
\label{novel_prot_result}
\centering \advance \leftskip-0.5cm
\begin{tabular}{lcccc}
\toprule
Model             & CI $\uparrow$                   & SPC $\uparrow$                  & PSC $\uparrow$                  & MSE $\downarrow$                  \\ \midrule
DeepconvDTI       & 0.52($\pm$0.02) & 0.17($\pm$0.09) & 0.17($\pm$0.08) & 2.89($\pm$0.87) \\
GraphDTA          & 0.54($\pm$0.05) & 0.10($\pm$0.16) & 0.11($\pm$0.15) & 4.28($\pm$3.09) \\
HyperAttentionDTI & 0.53($\pm$0.04) & 0.10($\pm$0.12) & 0.15($\pm$0.11) & 2.75($\pm$0.04) \\
PerceiverCPI      & 0.58($\pm$0.02) & 0.26($\pm$0.04) & 0.25($\pm$0.05) & \color{teal}2.55($\pm$0.40) \\
MulinforCPI       & \color{teal}0.61($\pm$0.04) & \color{purple}{0.34($\pm$0.11)} & \color{purple}{0.35($\pm$0.11)} & \color{purple}{2.14($\pm$0.24)} \\ \midrule
EquiCPI (ours)           & \color{purple}{0.61($\pm$0.03)} & \color{teal}0.32($\pm$0.10) & \color{teal}0.33($\pm$0.09) & 3.22($\pm$1.06) \\ \midrule
\end{tabular}
\end{table}

\begin{table}[h]
\centering
\caption{The decoy classification analysis results on a Diverse subset from the DUD-E database (EF$_{1\%}$ $\uparrow$ better, BEDROC$_{\alpha=80.5}$ $\uparrow$ better, mean and standard deviation values were computed from per protein results' averages, \color{purple}first performance, {\color{teal}second performance}, {\color{olive}third performance}.}
\begin{tabular}{lcc}
\toprule
Models      & EF$_{1\%}$ ($\pm$std) & \begin{tabular}[c]{@{}c@{}}BEDROC$_{\alpha=80.5}$\\ ($\pm$std) \end{tabular}
      \\\midrule
DeepConvDTI & 6.357($\pm$6.173)   & 0.118($\pm$0.109) \\
TransformerCPI   & 7.039($\pm$12.496)  & 0.117($\pm$0.192) \\
HyperattentionDTI  & 1.753($\pm$2.551)   & 0.038($\pm$0.051) \\
PerceiverCPI  & 4.649($\pm$3.136)   & 0.094($\pm$0.067)  \\
MulinforCPI   & \color{teal}7.886($\pm$10.642)  & 0.137($\pm$0.167) \\  \midrule 
EquiCPI (ours)    & \color{olive}7.820($\pm$8.864)  & \color{olive}0.142($\pm$0.145) \\ \midrule

Random Guessing & 0.940($\pm$0.844)   & 0.022($\pm$0.010) 
 \\\midrule
Gold  & N/A            & \color{teal}0.253($\pm$0.182) \\
Glide  & N/A            & \color{purple}{0.259($\pm$0.171)} \\
Surflex & N/A            & 0.119($\pm$0.093) \\
FlexX & N/A            & 0.104($\pm$0.060)  \\
Blaster & \color{purple}{13.571($\pm$12.908)} & N/A  \\\midrule 

\end{tabular}
\label{ef_ana}
\end{table}

EquiCPI consistently excels in the first three metrics C-index, Spearman correlation, and Pearson correlation, surpassing the competing models and establishing itself as the leading approach. This indicates that the model's predictions have a strong linear relationship with the actual values. Notably, in Figure \ref{novel_pair_result}, the hardest settings, EquiCPI achieves the highest Spearman correlation and highest Pearson correlation, reflecting that the predicted rankings closely match the actual rankings and values.

These results indicate that EquiCPI not only predicts values that closely align with actual observations in both magnitude and rank but also effectively captures intricate patterns, including both linear and non-linear relationships within the data. The model’s high correlation scores underscore its robustness and reliability, demonstrating its ability to generalize well across diverse datasets. While EquiCPI does not consistently outperform all competing models in every metric, particularly in terms of MSE (as shown in Tables \ref{novel_comp_result} and \ref{novel_prot_result}), it still exhibits strong overall performance. Its balanced predictive power across multiple evaluation criteria suggests that it remains a highly competitive model, providing meaningful insights while maintaining consistency across different benchmarks.

To evaluate EquiCPI’s performance in virtual screening benchmarks, we compare its early enrichment capacity, binding detection reliability, and stability against state-of-the-art machine learning (ML) approaches and classical docking tools. Table \ref{ef_ana} presents a consolidated view of these metrics. EquiCPI achieves an EF\textsubscript{1\%} score of 7.82 ± 8.86, closely matching MulinforCPI (7.89) while significantly outperforming baseline docking models such as Blaster (13.57). BEDROC applies an exponential weighting function (Boltzmann distribution) to prioritize early-ranked active compounds more than those ranked later. EquiCPI ranks second among ML-based approaches, attaining a BEDROC\textsubscript{80.5} score of 0.142 ± 0.145, surpassing MulinforCPI (0.137) and maintaining 56\% of Gold/Glide’s performance (0.253-0.259) without relying on explicit physics-based scoring. Furthermore, EquiCPI demonstrates greater consistency across diverse target classes, with a lower BEDROC\textsubscript{80.5} standard deviation (0.145) compared to MulinforCPI (0.167), suggesting a more stable performance profile. The variance in EF\textsubscript{1\%} (8.864) is also comparable to TransformerCPI (12.496), highlighting EquiCPI’s robustness. While it does not achieve absolute superiority in all metrics, EquiCPI offers a well-balanced trade-off between early enrichment, computational efficiency, and generalizability, making it a practical tool for real-world drug discovery workflows where early enrichment is critical for hit identification.

\subsection{Combining Physics-Based Docking Scores with DiffDock-L for Enhanced CPI Prediction}

\begin{figure}[h]
    \centering
    \includegraphics[scale=0.5]{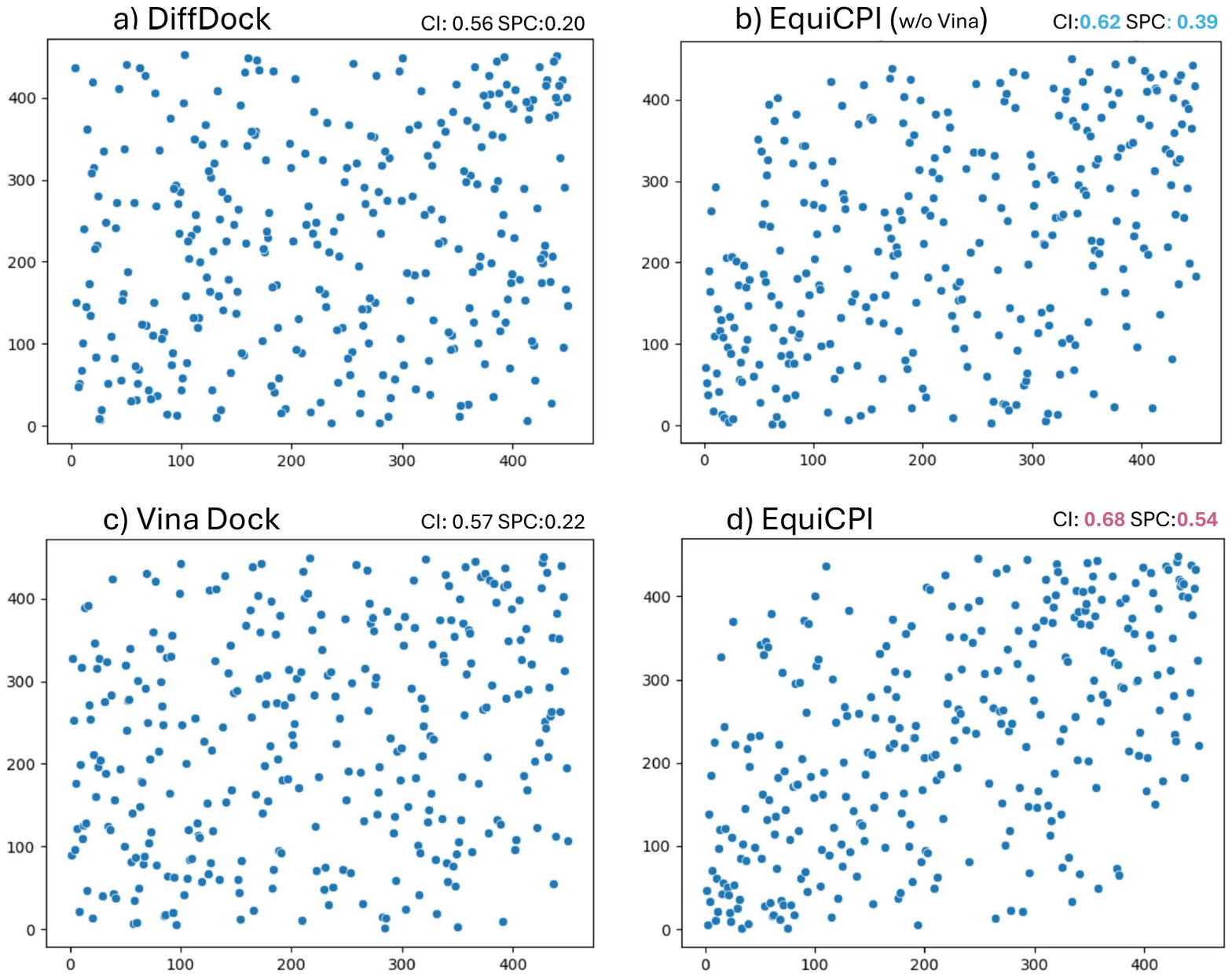}
     \caption{The combinations of deep learning techniques and docking techniques (500 testing points of Fold0 novel compound setting).}
    \label{docking_combines}
\end{figure}

\begin{figure}[h]
    \centering \advance \leftskip-1.4cm
    \includegraphics[scale=0.22]{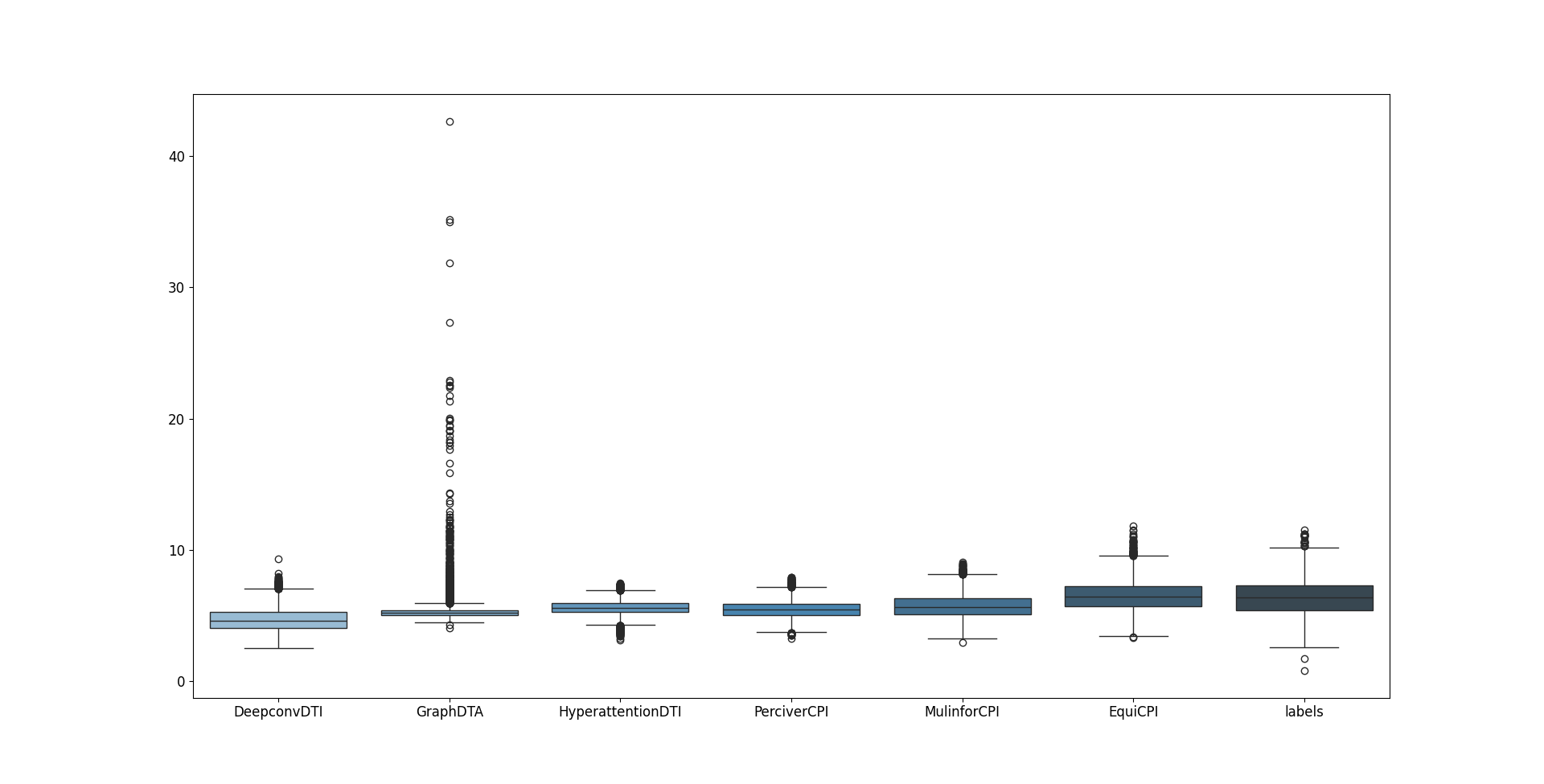}
     \caption{The distribution of predictions for EquiCPI and its competitors in the first fold of the novel compound setting.}
    \label{novel_pair_0}
\end{figure}

As illustrated in Figure \ref{docking_combines}, the combined approach of utilizing deep learning techniques for pose prediction alongside docking scores for pose ranking proves to be a superior method. By integrating these two strategies, we achieve more precise and accurate predictions of molecular poses for a binding affinity prediction task. Machine learning enhances the predictive capability by learning complex patterns, while docking scores provide a quantitative ranking of the predicted poses based on their energetic favorability. This hybrid method outperforms traditional approaches that rely solely on docking scores for both pose prediction and ranking, offering a more reliable and efficient framework for molecular docking analysis. The synergy between these techniques allows for improved accuracy in identifying the most biologically relevant poses, ultimately leading to more robust predictions and better overall performance.

Figure \ref{docking_combines} also highlights that when used independently, neither machine learning-based pose prediction nor docking score ranking alone performs as well as their combination. The standalone application of these techniques results in suboptimal outcomes, with each approach falling short in either predictive accuracy or ranking reliability. However, the combination of the two methods significantly improves performance. Notably, the figure shows that EquiCPI achieves the highest overall performance, outperforming both standalone techniques and other combined approaches. This emphasizes the advantage of leveraging EquiCPI for pose prediction and ranking, as it offers the most effective and accurate results in this study.

As demonstrated in Figures \ref{novel_pair_0}, the prediction distribution of EquiCPI aligns closely with the true label distribution, outperforming existing methods \citep{nguyen2023perceiver, lee2019deepconv, nguyen2021graphdta, zhao2022hyperattentiondti, nguyen2024mulinforcpi} in capturing the statistical patterns of compound-protein interactions. This robust agreement underscores the model’s capacity to generalize beyond training data, suggesting it effectively learns the underlying biological mechanisms governing these interactions. Unlike benchmarked approaches, EquiCPI preserves the structural and functional dependencies between compounds and proteins, which is critical for maintaining predictive fidelity in complex biological systems. The observed consistency between predicted and experimental distributions further validates that our framework successfully disentangles and retains biologically relevant features—such as binding affinity determinants and interaction hotspots—that are essential for advancing prediction accuracy in drug discovery. This capability positions EquiCPI as a promising tool for modeling high-dimensional, non-linear relationships in chemoproteomic studies.

\subsection{Discussion}
Although many studies have individually employed structural predictions, physics-based scoring, or SE(3)-equivariant networks—for example, using EquiBind for docking or NequIP for interatomic potentials—this paper uniquely combines these approaches into a unified, end-to-end framework for compound–protein interaction prediction. The method leverages state-of-the-art structure prediction tools (ESMFold and DiffDock-L) to generate accurate 3D atomic coordinates, utilizes physics-based conformer re-ranking via AutoDock Vina, and then applies SE(3)-equivariant message passing over heterogeneous graphs to predict binding affinities. This seamless integration—bridging first-principles modeling with geometric deep learning tailored to capture critical 3D spatial determinants—addresses the limitations of sequence-based and 2D graph-based methods, marking a significant advancement in computational drug discovery. While our hybrid framework significantly advances compound–protein interaction prediction by combining 3D structural modeling with equivariant deep learning, several methodological and practical challenges still warrant careful consideration.

\textbf{Structural Prediction Uncertainties}: The accuracy of downstream binding affinity predictions remains contingent on the fidelity of initial 3D structure generation. For ligands, DiffDock-L’s ($\approx$43 top-1 success rate (RMSD < 2\r{A}) implies that 57 of the predicted poses require energy minimization or experimental refinement before reliable feature extraction. Similarly, ESMFold’s local structure errors in flexible loop regions (Local Distance Difference Test(lDDT) < 60) may misrepresent critical binding interfaces, particularly for allosteric sites. These uncertainties propagate through the pipeline, as equivariant networks implicitly assume structural correctness during geometric feature extraction.

\textbf{Computational and Input Constraints}: The sequential execution of DiffDock-L ($\approx$ 10s/ligand) and ESMFold ($\approx$ 15s/protein) creates bottlenecks for large-scale virtual screening. While faster than traditional MD simulations, this remains impractical for billion-compound libraries compared to classical QSAR methods. Furthermore, the framework inherits limitations of its input representations: SMILES ambiguity (tautomerism, stereochemistry) and ESMFold’s inability to model non-canonical residues or post-translational modifications restrict applicability to idealized biological systems.

\textbf{Equivariance-Accuracy Trade-offs}: Though SE(3)-equivariant architectures theoretically preserve geometric relationships, their rigid-body transformation assumptions neglect quantum mechanical effects critical for precise binding energy calculations. This manifests as systematic errors in predicting interactions involving metal-coordination bonds. The current implementation also fails to account for entropic contributions or solvent effects, unlike free energy perturbation approaches.

\textbf{Generalization Challenges}: Performance degradation occurs for:
\begin{itemize}
    \item Novel chemotypes: DiffDock-L fails to accurately predict binding poses for most macrocyclic compounds and covalent inhibitors due to their complex binding mechanisms. (DiffDock-L success rate < 20\%).
    \item Long sequence proteins: ESMFold struggles with long protein sequences, often predicting structures with low accuracy (TM-score < 0.5).
\end{itemize}

\textbf{Disordered regions}: Intrinsically disordered protein interfaces.

These limitations mirror broader challenges in structural bioinformatics but are exacerbated by our pipeline’s dependency on predicted rather than experimental structures.

\textbf{Comparative Context}: Compared to traditional docking, our framework improves binding affinity prediction accuracy while introducing new computational overheads. The equivariant network’s geometric awareness provides advantages over invariant GNNs but at the cost of interpretability.

\section{Conclusion and Future Directions}

In summary, our work presents a unified, end-to-end framework that integrates cutting-edge structure prediction (using ESMFold and DiffDock-L), physics-based conformer re-ranking (via AutoDock Vina), and SE(3)-equivariant message passing over heterogeneous graphs to predict compound-protein interactions. This integration not only ensures that our predictions remain consistent under geometric transformations but also leverages the full three-dimensional information of the molecular structures, overcoming limitations of traditional sequence-based and 2D graph-based approaches.

Our key contributions are:
\begin{itemize}
    \item The novel integration of 3D structural modeling with physics-based re-ranking and SE(3)-equivariant deep learning.
    \item A robust method that captures the intrinsic 3D spatial determinants of binding affinity.
    \item Extensive validation on benchmark datasets that confirms the model’s superior performance and generalizability.
\end{itemize}

Looking forward, future research can focus on further reducing computational complexity and exploring the integration of additional biophysical constraints to enhance model interpretability. Moreover, prospective experimental validation and application to real-world drug screening scenarios are promising directions to establish the practical impact of our framework. We believe that our approach paves the way for more efficient and accurate computational drug discovery, bridging the gap between theoretical advances and practical applications. We propose potential future directions.

\begin{itemize}
    \item Uncertainty-aware modeling: Bayesian neural networks to quantify prediction confidence \citep{kyro2025t}.
    \item Hybrid physics-ML scoring: Integrate MM/GBSA energy terms with learned features \citep{roberto2025investigation,amin2025towards}.
    \item Active learning pipelines: Prioritize high-entropy predictions for experimental validation.
    \item Extended structural sampling: Incorporate explicit solvent and side-chain rotamers.
\end{itemize}

\section*{Funding}
This work was supported in part by the National Research Foundation of Korea [NRF-2023R1A2C3004176], the Ministry of Health \& Welfare, Republic of Korea [HR20C0021(3)], the Ministry of Science and ICT (MSIT) [RS-2023-00262002], and the ICT Creative Consilience program through the Institute of Information \& Communications Technology Planning \& Evaluation(IITP) grant funded by the MSIT [IITP-2024- 2020-0-01819]. This work was also supported by the Korea Health Technology R\&D Project grant through the Korea Health Industry Development Institute (KHIDI), funded by the Ministry of Health \& Welfare, Republic of Korea [HR22C1302].

\section*{Data Availability}
The source data, network codes can be found on GitHub at:\\
\url{https://github.com/dmis-lab/EquiCPI}
\begin{suppinfo}

A list of the contents of each file provided as Supporting Information. The following files are available free of charge.
\begin{itemize}
  \item JCIM\_equiCPI\_sup.pdf: Supplemental Information

\end{itemize}

\end{suppinfo}

\bibliographystyle{natbib}
\bibliography{achemso-demo}

\end{document}